\documentclass[sigconf]{acmart}
\usepackage{subfig}
\usepackage{soul}
\usepackage{url}
\usepackage[utf8]{inputenc}

\usepackage{amssymb}
\usepackage{amsthm}
\usepackage{pifont}
\usepackage{threeparttable}
\usepackage{balance}
\AtBeginDocument{%
  }
    
\setcopyright{acmlicensed}
\copyrightyear{2025}
\acmYear{2025}
\acmDOI{3746027.3755334}
\acmConference[MM '25]{the 33rd ACM International Conference on Multimedia}{October 27--31, 2025}{Dunlin, Ireland}
\acmISBN{979-8-4007-2035-2/25/10}

\begin{document}

\title{GTHNA: Local-global Graph Transformer with Memory Reconstruction for Holistic Node Anomaly Evaluation}

\author{Mingkang Li}
\authornote{Both authors contributed equally to this research.}
\orcid{0009-0005-1099-1914}

\affiliation{%
  \department{School of Computer Science}
  \institution{Wuhan University}
  \city{Wuhan}
  \state{Hubei}
  \country{China}
}
\email{mingkangli@whu.edu.cn}

\author{Xuexiong Luo}
\authornotemark[1]
\orcid{0000-0003-3400-4061}
\affiliation{%
  \department{School of Computing}
  \institution{Macquarie University}
  \city{Sydney}
  \state{NSW}
  \country{Australia}
}
\email{xuexiong.luo@hdr.mq.edu.au}

\author{Yue Zhang}
\orcid{0009-0008-6101-260X}
\affiliation{%
  \department{School of Cyber Science and Engineering}
  \institution{Wuhan University}
  \city{Wuhan}
  \state{Hubei}
  \country{China}
}
\email{yuezhang@whu.edu.cn}

\author{Yaoyang Li}
\orcid{0009-0009-0535-4928}
\affiliation{%
  \institution{Wuhan University}
  \city{Wuhan}
  \state{Hubei}
  \country{China}
}
\email{liyaoyang@whu.edu.cn}

\author{Fu Lin}
\authornote{Corresponding author.}

\orcid{0009-0002-3749-4043}

\affiliation{%
  \department{School of Computer Science}
  \institution{Wuhan University}
  \city{Wuhan}
  \state{Hubei}
  \country{China}
}
\email{linfu@whu.edu.cn}
\renewcommand{\shortauthors}{Mingkang Li, Xuexiong Luo, Yue Zhang, Yaoyang Li, and Fu Lin}

\begin{abstract}
Anomaly detection in graph-structured data is an inherently challenging problem, as it requires the identification of rare nodes that deviate from the majority in both their structural and behavioral characteristics. Existing methods, such as those based on graph convolutional networks (GCNs), often suffer from over-smoothing, which causes the learned node representations to become indistinguishable. Furthermore, graph reconstruction-based approaches are vulnerable to anomalous node interference during the reconstruction process, leading to inaccurate anomaly detection. In this work, we propose a novel and holistic anomaly evaluation framework that integrates three key components: a local-global Transformer encoder, a memory-guided reconstruction mechanism and a multi-scale representation matching strategy. These components work synergistically to enhance the model’s ability to capture both local and global structural dependencies, suppress the influence of anomalous nodes, and assess anomalies from multiple levels of granularity. Anomaly scores are computed by combining reconstruction errors and memory matching signals, resulting in a more robust evaluation. Extensive experiments on seven benchmark datasets demonstrate that our method outperforms existing state-of-the-art approaches, offering a comprehensive and generalizable solution for anomaly detection across various graph domains.
\end{abstract}

\ccsdesc[500]{Computing methodologies~ Artificial intelligence}
\ccsdesc[500]{Learning paradigms~Graph-based learning}

\keywords{Unsupervised graph anomaly detection, Graph Transformer, Graph representation learning}


\maketitle

\section{Introduction}
Graph-structured data is essential for capturing and analyzing complex relationships in real-world scenarios, including computer networks and social networks \cite{velivckovic2023everything}. The powerful expressive capability of graph data has led to its widespread application in network analysis\cite{kumar2019predicting} and recommendation systems\cite{ma2021comprehensive}. Hence, how to find anomalous patterns or outliers in graph-structured data has become a crucial research area. Anomaly detection in graph-structured data aims to identify rare entities that deviate from the majority of nodes, and these entities may exhibit anomalous behavior. This capability is critical in applications such as network attack detection \cite{wu2020network}, financial fraud prevention \cite{kurshan2020financial}, and social spam detection \cite{rao2021review}. Despite its importance, it continues to face significant challenges.

\begin{figure}[tbp] 
    \centering
    \includegraphics[width=0.48\textwidth ]{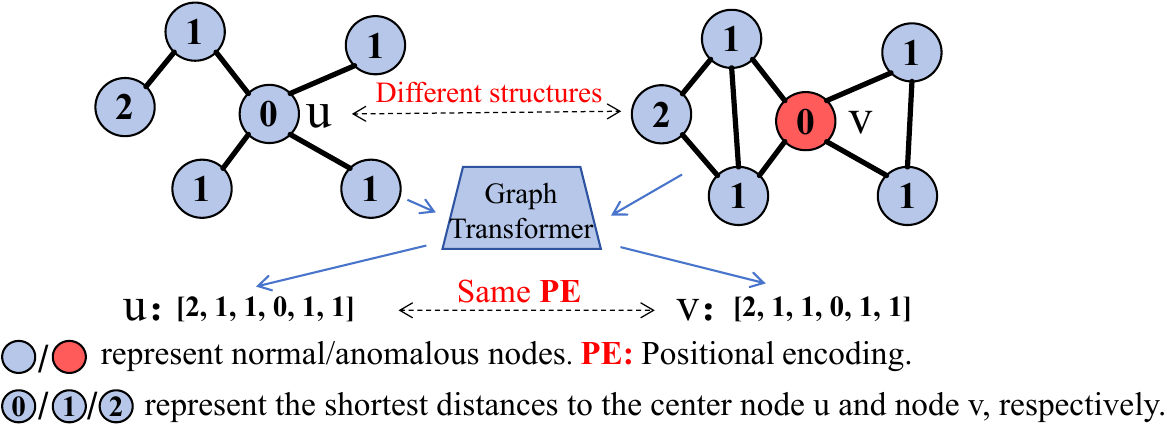} 
    \caption{Normal node $u$ and anomalous node $v$ have distinct topological structures. However, when encoded by a graph Transformer using absolute positional encoding (shortest path), they get same positional encoding. This erases structural differences between nodes, weakening the model's ability to detect structure anomalies at the node level.}
    \label{fig:Figure 1} 
\end{figure}

The complexity of graph data poses challenges for extracting rich and effective node representations, as these representations should be sensitive to their own structure and attribute, while also being distinguishable \cite{tang2022rethinking}. Current methods typically rely on graph convolutional networks (GCNs) to learn node representations. However, GCNs are prone to the issue of over-smoothing, where node features become indistinguishable after multiple layers of message passing \cite{chen2020measuring}. The graph Transformer \cite{wu2024demystifying} is proposed to address the over-smoothing issue by incorporating global graph information. TransGAD \cite{guo2024transgad} is the first to apply graph Transformer and introduces the cosine positional encoding methods. Despite its advantages, it relies on an absolute positional encoding method. This method computes positional encodings based solely on node connections, without incorporating contextual information from the neighborhood \cite{shaw2018self}. \figurename~\ref{fig:Figure 1} shows that the graph Transformer with absolute positional encoding may fail to capture the structural differences between nodes. As a result, it becomes challenging to distinguish anomalous nodes from normal ones, even though they have distinct topologies. This limitation highlights the need for graph Transformer that can accurately capture both local and global graph structure's information while mitigating the over-smoothing problem.

Moreover, the lack of labeled anomalies in graph data limits the applicability of supervised learning approaches. As a result, most existing methods adopt unsupervised or semi-supervised techniques. Apart from graph contrastive learning-based methods \cite{lu2024enhancing,zheng2021generative}, most approaches \cite{fan2021hyperspectral,du2022graph} are based on graph reconstruction and use node reconstruction errors as the basis for determining whether a node is anomalous. For example, DOMINANT \cite{ding2019deep} trains model to accurately reconstruct graph structure and node attributes. Nodes are then evaluated based on their reconstruction errors: higher errors are indicative of anomalies. 

\begin{figure}
    \centering
    \subfloat[]{
       \includegraphics[width=0.45\linewidth]{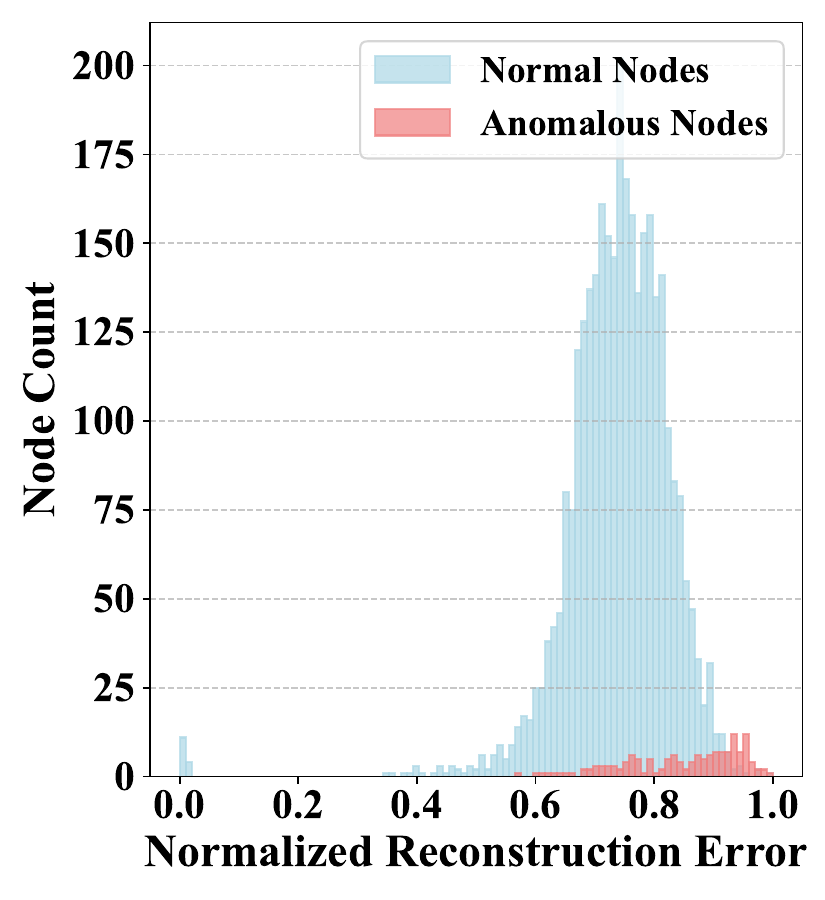}\hfill 
    }
    \hfill 
    \subfloat[]{
       \includegraphics[width=0.45\linewidth]{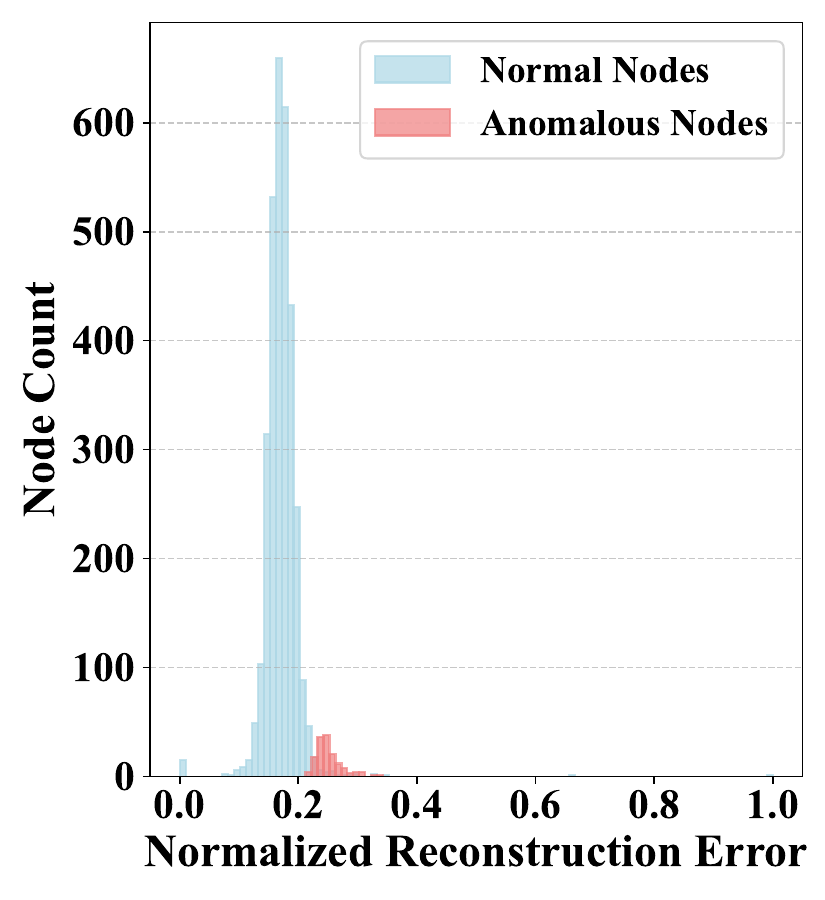}\hfill
    }
    \caption{The normalized reconstruction error distribution generated by DOMINANT (a) and our proposed method (b) on Citeseer dataset: In (a), the red and blue areas overlap significantly, whereas in (b), the overlap is less, indicating that some anomalous nodes' reconstruction errors in (a) are closer to normal nodes, making it more difficult for the model to distinguish these anomalous nodes. }
    \label{fig:Figure 2}
\end{figure}

However, graph reconstruction-based approaches commonly operate under the assumption that all nodes are normal during training, aiming to minimize reconstruction error. As illustrated in \figurename~\ref{fig:Figure 2}, this assumption may result in certain anomalous nodes—particularly those that closely resemble normal nodes—being reconstructed with low error, thereby making them harder to detect. Consequently, the model's ability to distinguish between normal and anomalous nodes is compromised, which limits its overall performance in anomaly detection. Meanwhile, typical graph reconstruction-based methods primarily restore the structural information by directly reconstructing links between nodes, while neglecting the information from the entire neighborhood. This restricts the model's ability to understand the complex graph structure \cite{roy2024gad}.

To address the above problems, we propose a holistic anomaly evaluation framework that unifies three core components, which termed Local-global \textbf{G}raph \textbf{T}ransformer with Memory Reconstruction for \textbf{H}olistic \textbf{N}ode \textbf{A}nomaly Evaluation (GTHNA\footnote{The source code is available at \url{https://github.com/light-flow/GTHNA}}). First, a local-global graph Transformer encoder is designed to capture both local and global graph context, ensuring that node representations are distinguishable and robust against over-smoothing. Second, a memory-guided reconstruction mechanism is introduced to store prototypical patterns of normal nodes, guiding the reconstruction process and suppressing the influence of anomalous nodes. Third, we leverage a multi-scale anomaly evaluation strategy to capture anomalies at different structural and attribute granularities. By assessing node behaviors across multiple neighborhood scopes and feature resolutions, the model can better distinguish subtle anomalies that may be overlooked when using a single-scale perspective. The final anomaly scores are computed by combining graph reconstruction errors and memory matching degrees, enabling a more comprehensive evaluation of anomalies. Our main contributions are summarized as follows:
\begin{itemize}
    \item [$\bullet$]\textbf{Local-global Graph Transformer Encoder:} We introduce a local-global graph Transformer encoder that captures both local and global structural information. This makes the node representations of normal and anomalous nodes more distinguishable.
    \item [$\bullet$]\textbf{Memory-guided Graph Reconstruction:} In the graph reconstruction process, we use a memory module to store the patterns of normal nodes and guide the node reconstruction with memory items, ensuring more accurate reconstruction of normal nodes and reducing the interference of anomalous nodes in the reconstruction process.
    \item [$\bullet$]\textbf{Multi-scale Anomaly Evaluation:} By incorporating the multi-scale reconstruction mechanism, we enhance the model's ability to capture complex graph structures at various levels, improving the robustness of anomaly detection.
\end{itemize}

\begin{table*}[!th]
\centering
\scalebox{0.95} {
\begin{threeparttable}
\begin{tabular}{cccccc}
\toprule
Method & Graph Encoder & Structure Mining & Over-Smoothing Mitigation & Reconstruction Way & Anomaly Evaluation \\ \midrule
DOMINANT \cite{ding2019deep}       & GCN  & Local & \ding{55}  & Standard &  RE  \\
AnomalyDAE \cite{fan2020anomalydae}& GAT  & Local &\ding{55} & Standard & RE \\
VGOD \cite{huang2023unsupervised}  & GAT  & Global &\ding{55} &  Standard &  RE \\
GAD-NR \cite{roy2024gad}           & GCN  & Local  & \ding{55} &  Multi-scale &  RE \\
TransGAD \cite{guo2024transgad}    & GT   & Global & \ding{51} & Standard & RE \\ \midrule
\textbf{GTHNA(Ours)}    & \textbf{GT}  &\textbf{Global \&  Local}  & \ding{51} & \textbf{Multi-scale} &  \textbf{MMD  \& RE} \\ \bottomrule
\end{tabular}
\begin{tablenotes}[para,flushleft]
Notes: \textbf{GCN} is short for Graph Convolutional Network. \textbf{GAT} is short for Graph Attention Network. \textbf{GT} is short for Graph Transformer. \textbf{RE} is short for Reconstruction Error. \textbf{MMD} is short for Memory Matching Degree.
\end{tablenotes}
\end{threeparttable}
}
\caption{Comparison with mainstream graph reconstruction-based anomaly detection methods.}
\label{tab:1}
\end{table*}
\section{Related Work}
\subsection{Graph Anomaly Detection}
Due to the rarity of graph anomalies and the difficulty of labeling, anomaly detection methods primarily adopt unsupervised or semi-supervised approaches. Broadly, we categorize these methods into two types: graph contrastive learning-based methods \cite{liu2021anomaly,jin2021anemone} and graph reconstruction based methods \cite{ding2019deep,roy2024gad}. Graph contrastive learning-based methods create augmented views of the graph and maximize the similarity of a node's embeddings across views (positive pairs) while minimizing the similarity with other nodes (negative pairs). Anomalies are identified based on inconsistencies in node embeddings across views. 

However, graph contrastive learning-based methods are highly sensitive to data augmentation strategies, as poorly designed augmentations may disrupt critical graph structures. Different from graph contrastive learning-based methods, graph reconstruction-based methods aim to learn the information from the original graph directly. They provide more interpretable signals for identifying anomalies, as high reconstruction errors indicate specific deviations in structure or features. DOMINANT \cite{ding2019deep} first introduces graph reconstruction into graph anomaly detection. It consists of a shared graph encoder and two kinds of decoders which are designed for reconstructing structure and attributes respectively. Then, AnomalyDAE \cite{fan2020anomalydae} uses a dual autoencoder to capture the complex interactions between graph structure and node attributes, ComGA \cite{luo2022comga} leverages community information of nodes to make a more targeted distinction between normal and anomalous nodes, VGOD \cite{huang2023unsupervised} combines variance-based model and attribute model to handle the issue of balanced detection, and GAD-NR \cite{roy2024gad} adds the reconstruction of node neighborhood to improve the model's anomaly detection performance. 

In \tableautorefname~\ref{tab:1}, we compare our work with previous mainstream graph reconstruction-based methods and show that all prior methods do not fully explore graph structure, as they typically focus on either local or global structure mining. Additionally, they rely solely on reconstruction error for node anomaly evaluation. These problems limit their anomaly detection performance. Compared with existing approaches that rely solely on reconstruction error and often overlook the structural diversity of anomalies, our proposed GTHNA framework introduces a fundamentally different perspective. Rather than treating anomaly detection as a single-stage task, GTHNA establishes a memory-aware, context-rich, and multi-resolution pipeline that systematically enhances each phase of the process—from representation learning to evaluation. By explicitly modeling normal patterns through the memory module and integrating local-global context via Transformer-based encoding, our framework ensures that the anomaly detection process is both discriminative and resilient to noise or subtle perturbations. The holistic design of GTHNA not only improves detection performance across various types of graph anomalies but also demonstrates strong generalization capability across different domains. 
\subsection{Graph Transformer}
The graph Transformer is proposed primarily to address the over-smoothing problem encountered in graph convolutional processes. By incorporating global information, the graph Transformer makes node representations can focus on broader context, not be limited by the feature propagation from the neighborhood. But the typical graph Transformers lack effective methods for utilizing structure information. Subsequently, many improved graph Transformers have emerged. Graph-BERT \cite{zhang2020graph} employs multiple types of positional encoding methods and skip connections to introduce the structure information into the node embeddings. EGT \cite{hussain2021edge} proposes residual edge channels to evolve structure information from layer to layer. But most of these graph Transformers take absolute positional encodings solely from the graph's adjacency matrix and are not sufficient to fully capture the structural differences between nodes. \cite{chen2022structure}.

On the other hand, although graph Transformers have become increasingly mature, how to apply them to graph anomaly detection has not been thoroughly investigated. Though TransGAD \cite{guo2024transgad} attempts to use the graph Transformer as encoder and alleviates the over-smoothing issue, its exploration of the graph structure remains insufficient, as anomalies in graph may arise from subtle deviations in node connectivity or feature distribution that requires a deep understanding of both local and global graph structure. To address this challenge, we design a local-global graph Transformer, which explicitly captures both fine-grained local neighborhood structures and high-level global dependencies. This dual-perspective encoding enables the model to detect anomalies arising from either localized perturbations or global inconsistencies. Furthermore, we integrate a memory-guided graph process into the Transformer architecture, allowing the model to retain prototypical patterns of normal nodes and suppress the impact of anomalous information during representation learning and graph reconstruction. This design not only enhances the model’s sensitivity to subtle deviations but also improves its robustness and interpretability in detecting diverse types of anomalies.
\section{Preliminaries}
In this section, we present the definition of the graph and the graph anomaly detection task.
\begin{definition}
\textbf{Graph.} We primarily study the undirected attribute graph that can be represented as $\mathcal{G} = (\mathcal{V}, \mathcal{E}, \mathbf{\mathcal{X}})$, where $\mathcal{V} = \{v_1, v_2, ... , v_n\}$ represents the node set with $|\mathcal{V}| = n$, $\mathcal{E}$ denotes the set of edges connecting the nodes, $\mathbf{\mathcal{X}} = \{\mathbf{x}_1, \mathbf{x}_1, ..., \mathbf{x}_n\}$, $\mathbf{\mathcal{X}} \in \mathbb{R}^{n \times d}$ represents the node feature matrix of $\mathcal{G}$. We use $\mathbf{\mathcal{A}} \in \mathbb{R}^{n \times n}$ to denote the adjacency matrix of $\mathcal{G}$, where $\mathbf{\mathcal{A}}_{ij} = 1$ if $(v_i, v_j) \in \mathcal{E}$ else $\mathbf{\mathcal{A}}_{ij} = 0$.
\end{definition}
\begin{definition}
\textbf{Graph Anomaly Detection.} Given an attribute graph $\mathcal{G}$,  we define the normal nodes set as $\mathcal{V}_0$ and the anomalous nodes set as $\mathcal{V}_1$. The goal of our framework is to learn a score function $Score(\cdot)$ to compute the anomaly score $s_i = Score(v_i)$ for each node $v_i \in \mathcal{V} = \mathcal{V}_0 \cup \mathcal{V}_1$. If $v_i \in \mathcal{V}_0$ and $v_j \in \mathcal{V}_1$, there should have $s_i < s_j$.
\end{definition}
\begin{figure*}[htbp] 
    \centering
    \includegraphics[width=1.0\textwidth]{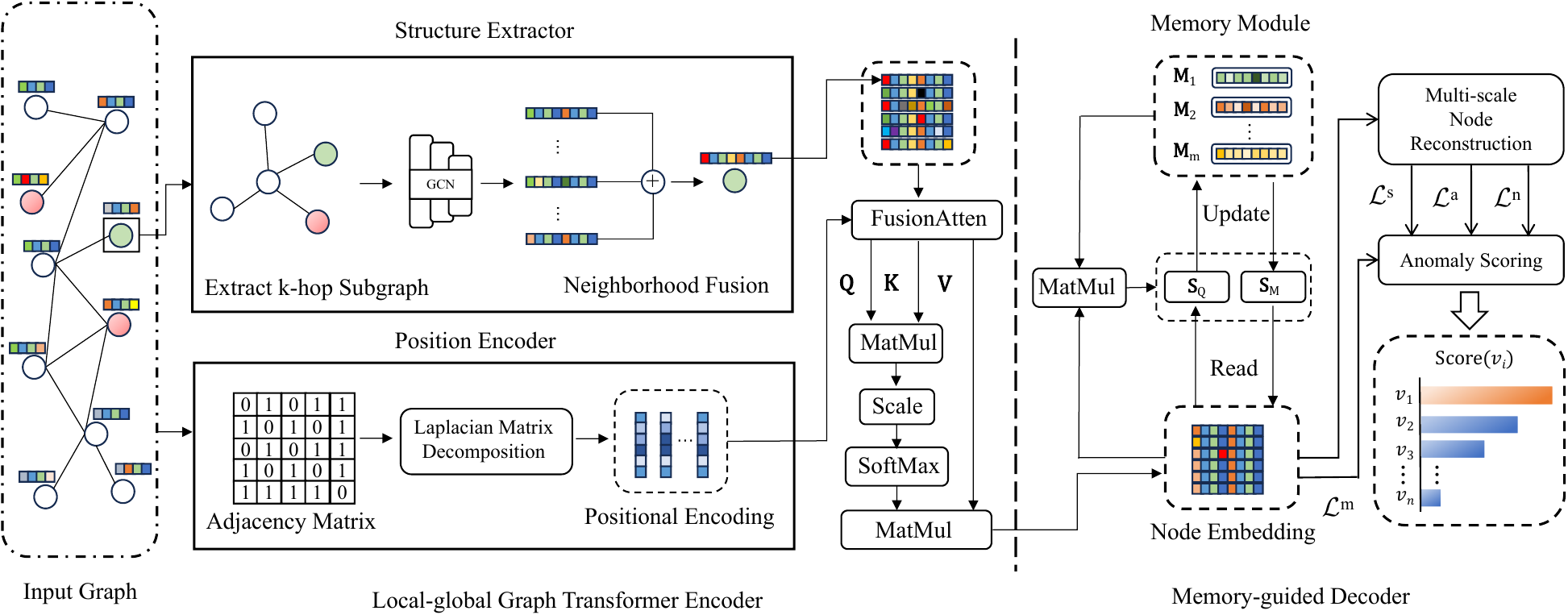} 
    \caption{The overview of GTHNA. The input nodes will be encoded by the local-global graph Transformer to get discriminate node embeddings and then decoded by the memory-guided decoder, we also introduce the multi-scale graph reconstruct strategy to capture node anomalies at different scales. The final node anomaly scores are computed based on the node reconstruction errors and the memory matching degrees.}
    \label{fig:Figure 3} 
\end{figure*}
\section{Methodology}
In this section, we delve into the details of GTHNA. As illustrated in \figurename~\ref{fig:Figure 3}, we use local-global graph Transformer to obtain node embeddings that incorporate both local and global graph information. Then, by leveraging the memory module that stores the normal node patterns, the model can focus more on reconstructing normal nodes. We introduce the multi-scale evaluation strategy to further improve the model's ability to capture anomalies at different graph scales. The final anomalous scores are computed by the reconstruction errors at different scales and the memory matching degrees.

\subsection{Local-global Graph Transformer}
\noindent \textbf{Structure Extractor.} To enhance the ability to capture local node structure information and obtain topologically sensitive node representations, we first extract the whole k-hop subgraph of the node $v_i$ and leverage GCN \cite{kipf2016semi} to fuse the neighborhood information of the node before performing a series of operations in the graph Transformer. This can be represented as:
\begin{equation}
    \mathbf{h}_i^{sub} = GCN^{(k)}(\mathcal{G}_i^{sub}, v_i),
\end{equation}
where $\mathcal{G}_i^{sub}$ is the k-hop subgraph rooted at node $v_i$, $GCN^{(k)}(\cdot, \cdot)$ is the GCN model with $k$ layers applied to $\mathcal{G}_i^{sub}$ using node features, and $\mathbf{h}_i^{sub} \in \mathbb{R}^{d_h}$ is the subgraph representation that captures the local graph information. Then, we can get $\mathbf{H}^{sub} = \{\mathbf{h}_1^{sub}, \mathbf{h}_2^{sub}, ..., \mathbf{h}_n^{sub}\}$ that denotes the learned node representations. In practice, a small value of $k$ can achieve strong performance without suffering from issues like over-smoothing or over-squashing.

\noindent \textbf{Position Encoder.} In typical graph Transformers, positional encoding is the primary approach of identifying node position information. In this paper, we use the Laplacian matrix to generate an absolute positional encoding for each node. The normalized Laplacian matrix $\mathbf{\hat{L}}$ of \(\mathcal{G}\) can be computed as:
\begin{equation}
    \mathbf{\hat{L}} = \mathbf{I} - \mathbf{D}^{-\frac{1}{2}}\mathbf{\mathcal{A}} \mathbf{D}^{-\frac{1}{2}},
\end{equation}
where $\mathbf{I}$ is the identity matrix, $\mathbf{D}$ is the degree matrix. Then the positional encoding $\mathbf{h}^{pos}_i$ of node $v_i$ is calculated by  performing eigenvalue decomposition on $\mathbf{\hat{L}}$:
\begin{equation}
    \mathbf{\hat{L}}\mathbf{\mu}_{(i)} = \mathbf{e}_i\mathbf{\mu}_{(i)}, \quad \mathbf{h}^{pos}_i =  \mathbf{\mu}_{(i)}^{[1, d_h]},
\end{equation}
here, $\mathbf{\mu}_{(i)}$ is the i-th eigenvector corresponding to eigenvalue $\mathbf{e}_i$ and $d_h$ is the dimension of node embeddings.

Although the positional encodings generated in this way can partially represent the positional characteristics of nodes, as previously discussed, relying solely on absolute positional encoding method is insufficient to fully capture the structure information \cite{chen2022structure}. Therefore, we combine the absolute positional encoding $\mathbf{h}^{pos}_i$ with the subgraph representation $\mathbf{h}^{sub}_i$ to obtain node representation before feeding it into the main Transformer model.
\begin{equation}
    \alpha_i = \sigma(\mathbf{W}^{(a)} \cdot [\mathbf{h}^{pos}_i, \mathbf{h}^{sub}_i]),
\end{equation}
\begin{equation}
    \mathbf{h}_i = \alpha_i \cdot \mathbf{h}^{pos}_i + (1 - \alpha_i) \cdot \mathbf{h}^{sub}_i,
\end{equation}
where $\alpha_i$ is the fusion attention weight computed as the sigmoid of the linear transformation $\mathbf{W}^{(a)}$ applied to the concatenation of $\mathbf{h}^{pos}_i$ and $\mathbf{h}^{sub}_i$. This approach enables the node representation $\mathbf{h}_i$ to incorporate both global structure information and local contextual information of node $v_i$ effectively.

\noindent \textbf{Graph Transformer-based Encoder.} After obtaining the node representation $\mathbf{h}_i$, we construct $\mathbf{H}_{\mathcal{G}} = \{\mathbf{h}_1, \mathbf{h}_2, ..., \mathbf{h}_n\}$ and use it as input to feed into the Transformer module. The Transformer consists of two main components: a self-attention mechanism followed by a feed-forward neural network. These operations can be formalized as follows:
\begin{equation}
    \mathbf{Q} = \mathbf{H}_{\mathcal{G}}\mathbf{W}^{(Q)}, \quad  \mathbf{K} =\mathbf{H}_{\mathcal{G}}\mathbf{W}^{(K)}, \quad \mathbf{V} = \mathbf{H}_{\mathcal{G}}\mathbf{W}^{(V)},
\end{equation}
\begin{equation}
    \mathbf{A} = \frac{\mathbf{QK}^\top}{\mathbf{\sqrt{d_K}}}, \quad \mathbf{H}_{\mathcal{G}}^{\prime} = softmax(\mathbf{A})\mathbf{V},
    \label{eq:1}
\end{equation}
where $\mathbf{Q}$, $\mathbf{K}$, and $\mathbf{V}$ represent the query, key, and value matrices, respectively; $d_K$ is the dimension of the key vector; $\mathbf{A}$ is the attention matrix; and $\mathbf{W}^{(Q)}, \mathbf{W}^{(K)}, \mathbf{W}^{(V)}$ are trainable weights. 

To improve model performance, multi-head attention (MHA) is commonly used, where multiple instances of \equationautorefname~\eqref{eq:1} are computed in parallel. The output is then followed by a skip connection and a feed-forward network (FFN), with layer normalization (LN) applied before each block, as shown below:
\begin{equation}
    \mathbf{H}_{\mathcal{G}}^{\prime} = MHA(LN(\mathbf{H}_{\mathcal{G}})) + \mathbf{H}_{\mathcal{G}},
\end{equation}
\begin{equation}
    \mathbf{H}_{\mathcal{G}}^{\prime \prime} = FFN(LN(    \mathbf{H}_{\mathcal{G}}^{\prime})) +\mathbf{H}_{\mathcal{G}}^{\prime},
\end{equation}
where $\mathbf{H}_{\mathcal{G}}^{\prime \prime} = \{\mathbf{h}_1^a, \mathbf{h}_2^a, ..., \mathbf{h}_n^a\}$ represents the node representations, each node representation $\mathbf{h}_i^a \in \mathbb{R}^{d_h}$, integrating both local and global graph information.
\subsection{Memory-guided Decoder}
\noindent \textbf{Memory Module.} Inspired by \cite{park2020learning}, we use the memory module to store $m$ items, each aims to represent a prototypical pattern of normal nodes. During unsupervised training, the memory module is dynamically updated by identifying nodes with relatively low reconstruction errors and treating them as pseudo-normal samples. These nodes are used to refine the memory items, encouraging the memory module to progressively capture representative patterns of normality and reduce the influence of anomalous nodes on the reconstruction process. We denote the memory items as $\mathbf{M} = \{\mathbf{M}_1, \mathbf{M}_2, ..., \mathbf{M}_m\}$ and the final node representations used for graph reconstruction are read from the memory module. As the model training, $\mathbf{M}$ is continuously updated.

To read the memory module, we use each node embedding vector $\mathbf{h}_i^a$ as the query vector, calculate the cosine similarity between it and each memory item $\mathbf{M}_j \in \mathbb{R}^{d_h}$, producing a correlation map of size $n \times m$. Next, we apply the softmax function along the horizontal axis to obtain the matching probabilities $\mathbf{S}_Q$, defined as:
\begin{equation}
    \mathbf{S}_{i, j}^Q  = \frac{exp(\mathbf{h}_i^a(\mathbf{M}_j)^\top)}{\sum_{j^\prime = 1}^{m} exp(\mathbf{h}_i^a(\mathbf{M}_{j^\prime})^\top)}.
    \label{eq:2}
\end{equation}
For each $\mathbf{h}_i^a$, the memory is read by taking a weighted average of the memory items, where the weights are given by $\mathbf{S}_Q$, The resulting node embedding $\mathbf{\hat{h}}_i \in \mathbb{R}^{d_h}$ is:
\begin{equation}
    \mathbf{\hat{h}}_i = \sum_{j^\prime = 1}^{m}  \mathbf{S}_{i, j^\prime}^Q \mathbf{M}_{j^\prime}.
\end{equation}
By leveraging all memory items, our model captures a diverse range of normal node characteristics. The concatenation of $\mathbf{h}_i^a$ and $\mathbf{\hat{h}}_i$, denoted as $\mathbf{\Bar{h}}_i \in \mathbb{R}^{2d_h}$, is then passed to the decoder. This allows the decoder to reconstruct the node based on the normal node patterns stored in the memory module, facilitating an understanding of normality.

To ensure that the memory module can store the most accurate patterns of normal nodes, we continuously update the memory module after each graph reconstruction. Using a process similar to \equationautorefname~\eqref{eq:2}, we apply the softmax function to the correlation map along a vertical axis to compute the matching probability $\mathbf{S}_M$ as: 
\begin{equation}
    \mathbf{S}_{j, i}^M = \frac{exp(\mathbf{h}_i^a(\mathbf{M}_j)^\top)}{\sum_{i^\prime = 1}^{n} exp(\mathbf{h}_{i^\prime}^a(\mathbf{M}_{j})^\top)}.
\end{equation}
Then we update the items using the query vectors as follows:
\begin{equation}
    \mathbf{M}_j \leftarrow \theta(\mathbf{M}_j+ \sum_{v_i \in U_h} \mathbf{S}_{j, i}^M \mathbf{h}_i^a),
\end{equation}
where $\theta(\cdot)$ denotes L2 normalization.
We use the nodes with smaller reconstruction errors, denoted as $U_h$, and assume that $U_h$ is the set of normal nodes, use it to update the memory items. These nodes are then used to update the memory items, guiding the memory module to focus on capturing representative patterns of normal behavior. By employing the weighted average of the node embeddings rather than a simple summation, this approach emphasizes the embeddings that are more closely aligned with the memory items.
\subsection{Multi-scale Node Reconstruction}
In order to capture multi-level graph structure information, we use $\mathbf{\Bar{h}}_i$ as input to perform multi-scale reconstruction for the node $v_i$.

\noindent\textbf{Node-self Reconstruction.} To restore the node attributes $\hat{\mathcal{X}}$, similar to the node encoding process, we use the graph Transformer as decoder to recover the node attributes as follows:
\begin{equation}
    \mathbf{\hat{\mathcal{X}}} = \mathcal{T}(\mathbf{\Bar{h}}_i|v_i \in \mathcal{V}),
\end{equation}
where $\mathcal{T}(\cdot)$ denotes the graph Transformer operation mentioned before. And we input the representations of two nodes to predict the existence of the edge $\hat{\mathcal{A}}_{i, j}$ between them:
\begin{equation}
    p(\hat{\mathcal{A}}_{i, j}| \Bar{\mathbf{h}}_i,\Bar{\mathbf{h}}_j) = \sigma(\Bar{\mathbf{h}}_i,\Bar{\mathbf{h}}_j^\top).
\end{equation}

\noindent \textbf{Node Neighborhood Reconstruction.} The actual neighborhood distribution $\mathcal{N}(\mathbf{\epsilon}_i, \mathbf{\omega}_i)$ of node $v_i$ can be represented as:
\begin{equation}
    \mathbf{\epsilon}_i = \frac{1}{c_i}\sum_{v_i \in \mathcal{H}_i}\mathbf{x}_i, \mathbf{\omega}_i = \frac{1}{c_i - 1}\sum_{v_i \in \mathcal{H}_i}(\mathbf{x}_i - \mathbf{\epsilon}_i)(\mathbf{x}_i - \mathbf{\epsilon}_i)^\top,
\end{equation}
where $\mathcal{H}_i$ denotes the one-hop neighbors set of node $v_i$ and $c_i$ is the count of its neighbors.

If the neighborhood information of a node can be accurately reconstructed from its embedding $\mathbf{\Bar{h}}_i$, it implies that the $\mathbf{\Bar{h}}_i$ contains multi-level graph structure information. However, directly reconstructing the information of each neighbor node is extremely challenging. Inspired by \cite{roy2024gad}, 
we indirectly reconstruct the neighborhood of a node by recovering both the count of its neighbors $\hat{c}_i$ and the distribution of neighbors' attributes $\mathcal{N}(\mathbf{\hat{\epsilon}}_i, \mathbf{\hat{\omega}}_i)$:
\begin{equation}
    \hat{c}_i = \varphi_c(\mathbf{\Bar{h}}_i), \mathbf{\hat{\epsilon}}_i = \varphi_\epsilon(\mathbf{\Bar{h}}_i), \mathbf{\hat{\omega}}_i = diag(exp(\varphi_\omega(\mathbf{\Bar{h}}_i))),
\end{equation}
where $\varphi_c(\cdot)$, $\varphi_\epsilon(\cdot)$ and $\varphi_\omega(\cdot)$ are FFN. 
\subsection{Training Loss and Anomaly Detection}
Following the assumption from \cite{ding2019deep} that a higher node reconstruction error indicates a greater likelihood of anomaly, we represent \(a_i \in \mathcal{A}\) as the adjacency relationship of node \(v_i\), \(\hat{a}_i \in \hat{\mathcal{A}}\) as the predicted adjacency vector and \(\hat{x}_i \in \hat{\mathcal{X}}\) as the reconstructed attribute vector. The multi-scale node reconstruction error is then calculated as follows:
\begin{equation}
    \mathcal{L}_i^s = \mathcal{D}(\mathbf{a}_i - \mathbf{\hat{a}}_i), \quad \mathcal{L}_i^a = \mathcal{D}(\mathbf{x}_i - \mathbf{\hat{x}}_i),
\end{equation}
\begin{equation}
    \mathcal{L}_i^n = KL(\mathcal{N}(\mathbf{\epsilon}_i, \mathbf{\omega}_i) || \mathcal{N}(\mathbf{\hat{\epsilon}}_i, \mathbf{\hat{\omega}}_i)) + \mathcal{D}(c_i - \hat{c}_i),
\end{equation}
where $\mathcal{D}(\cdot, \cdot)$ is a distance function such as L2-distance, KL is the Kullback-Leibler divergence between two distributions.

We also assume that the node embeddings from normal nodes will be similar to the memory items, which capture the prototypical patterns of normal nodes. The distance between each node embedding $\mathbf{h}_i^a$ and its nearest memory item $\mathbf{M}_f$ is computed as follows:
\begin{equation}
   \mathcal{L}_i^m = \mathcal{D}(\mathbf{h}_i^a, \mathbf{M}_f).
\end{equation}
To ensure memory module can store various patterns of normal nodes, the memory items should be sufficiently distinct from one another, we introduce a separateness loss, which is defined as:
\begin{equation}
    \mathcal{L}_i^p = ReLU(\mathcal{D}(\mathbf{h}_i^a, \mathbf{M}_f) -  \mathcal{D}(\mathbf{h}_i^a, \mathbf{M}_s) + \beta),
\end{equation}
where $\mathbf{M_f}$ is the nearest item, $\mathbf{M_s}$ is the second nearest item, and the $ReLU(\cdot)$ operation enforces the separation margin $\beta$. 

The overall training loss is then given by:
\begin{equation}
\begin{aligned}
 \mathcal{L} &= \sum_{v_i \in \mathcal{V}} [\lambda_s(\mathcal{L}_i^s + \mathcal{L}_i^a) + \lambda_n \mathcal{L}_i^a +\lambda_m (\mathcal{L}_i^m + \mathcal{L}_i^p)] \\
        &= \lambda_s(\mathcal{L}^s + \mathcal{L}^a) + \lambda_n \mathcal{L}^n + \lambda_m (\mathcal{L}^m + \mathcal{L}^p),
\end{aligned}
\label{eq:3}
\end{equation}
where $\lambda_s$, $ \lambda_n$ and $\lambda_m$ are hyperparameters that control the relative contributions of the node self-reconstruction loss, the node neighborhood reconstruction loss, and the memory module loss, respectively.

Finally, the anomaly score of node $v_i$ is computed based on the multi-scale node reconstruction error and the degree of matching between the node embedding and the memory items:
\begin{equation}
    Score(v_i) = \lambda_s(\mathcal{L}_i^s + \mathcal{L}_i^a) + \lambda_n \mathcal{L}_i^n +\lambda_m \mathcal{L}_i^m.
\end{equation}
\subsection{Complexity Analysis}
The overall time complexity of our proposed method is dominated by key components: 
\(\mathcal{O}(nd^2_h)\) for structure extractor where $n$ denotes node count and $d_h$ is the dimension of node embeddings, \(\mathcal{O}(n^3)\) for Laplacian matrix decomposition in positional encoding, \(\mathcal{O}(n^2d_h)\) for the graph Transformer encoder, \(\mathcal{O}(nm)\) for the memory module involving \(m\) memory items, and \(\mathcal{O}(|\mathcal{E}|)\) for the neighborhood reconstruction based on graph sparsity. Overall, the total complexity is approximately 
\(
\mathcal{O}(n^3 + nd^2_h + n^2d_h + nm + |\mathcal{E}|),
\)
where the relative importance of each term depends on the graph size, sparsity, and the dimension $d_h$ of node embeddings. It should be noticed that the Laplacian matrix decomposition can be completed before the training stage.

\section{Experiment}
In this section, we conduct a series of experiments. Specifically, we aim to answer the following questions:

\noindent\textbf{RQ1:} How does GTHNA perform compared to other graph anomaly detection methods?

\noindent\textbf{RQ2:} What are the impacts of the memory module and the structure extractor on model performance?

\noindent\textbf{RQ3:} Can the memory module effectively store the patterns of normal nodes?

\noindent\textbf{RQ4:} Can GTHNA make the anomaly scores between normal and anomalous nodes more distinguishable?

\noindent\textbf{RQ5:} How do key hyerparmeters such as $\lambda_s$, $\lambda_n$, and $\lambda_m$ influence the model performance?

\subsection{Experimental Setups}
\noindent\textbf{Datasets.} We choose seven benchmark datasets for evaluation. In particular, we organize these datasets into two categories: one consisting of datasets with artificially injected anomalies, including Citeseer \cite{sen2008collective}, ACM \cite{tang2008arnetminer}, and Blogcatalog \cite{tang2009relational}; and the other comprising datasets with organic anomalies, including Weibo \cite{zhao2020error}, Reddit \cite{kumar2019predicting}, YelpRes \cite{rayana2015collective}, and Amazon \cite{kaghazgaran2018combating}. The detailed statistics of these datasets are presented in \tableautorefname~\ref{tab:3}.
\begin{table}[h!]
\centering
\scalebox{1.0} {
\begin{tabular}{c|ccccc}
\toprule
Dataset     & \#Nodes & \#Edges & \#Features & \#Anomalies \\ \midrule
Citeseer    & 3,327   & 4,732   & 3,703          & 150         \\
ACM         & 16,484  & 71,980  & 8,337       & 600         \\
Blogcatalog & 5,196   & 171,743 & 8,189        & 300         \\ \midrule
Weibo       & 8,405   & 407,963 & 400         & 868         \\
Reddit      & 10,984  & 168,016 & 64         & 366         \\
YelpRes     & 5,012   & 355,144 & 8,000         & 250         \\
Amazon      & 17,496  & 495,213 & 10,000     & 705         \\ \bottomrule
\end{tabular}
}
\caption{Summary statistics for datasets used in experiment.}
\label{tab:3}
\end{table}

\noindent\textbf{Baselines.} We compare GTHNA with various types of open-source graph anomaly detection methods, including graph reconstruction-based methods (DOMINANT \cite{ding2019deep} and GAD-NR \cite{roy2024gad}), graph contrastive learning-based methods (CoLA \cite{liu2021anomaly} and ANEMONE \cite{jin2021anemone}), and graph augmentation-based methods (GADAM \cite{zhou2024graph}, SL-GAD \cite{zheng2021generative}, ADA-GAD \cite{he2024ada}, DiffGAD \cite{li2024diffgad}  and GRADATE \cite{duan2023graph}).

\noindent\textbf{Implementation Details.} In our experiment, the neighbor hop count \( k \) is set to 2, the node embedding dimension \( d_h \) is set to 128, and the Transformer encoder uses 4-head multi-head attention. The count of the memory items $m$ is set to 512. Since most of the nodes in the dataset are normal, the proportion of nodes with smaller reconstruction errors for $U_h$ can vary widely without significantly impacting model performance. In this work, it is set to 80\%. The margin \( \beta \) is set to 0.5, and the weights of the multi-scale reconstruction loss and memory module loss are set to \( \lambda_s = 1.0 \), \( \lambda_n = 0.01 \), and \( \lambda_m = 0.01 \). 

Additionally, we conduct experiments for both the baselines and GTHNA using 10 different random seeds (ranging from 0 to 9) and report the average and standard deviation of the results across these 10 runs as the final outcome.

\subsection{Performance Comparison (RQ1)}
We use the AUC to evaluate the anomaly detection performance of methods, as shown in \tableautorefname~\ref{tab:2}, We can observe that GTHNA outperforms almost all the baseline methods across all the datasets in detecting anomalies.

We can also observe that, on Citeseer and ACM, baselines based on graph augmentation and graph contrastive learning consistently outperform graph reconstruction-based baselines. In contrast, on Weibo, graph reconstruction-based baselines show superior performance. This can be attributed to the more apparent noise from anomalous nodes in injected datasets. Graph augmentation and contrastive learning techniques improve model robustness by introducing perturbations, enabling better handling of injected anomalies. But graph reconstruction-based baselines rely on the original graph structure to infer relationships, are more susceptible to the noise. On Weibo, the graph structure is free from artificial disturbances, allowing graph reconstruction-based methods to capture the real graph structure. In our approach, the local-global graph Transformer enhances the extraction of graph structure information, which make node embeddings be more discernible. Additionally, the memory module and can guide the graph reconstruction process, mitigating the impact of noise. The synergistic interaction between the encoder and decoder not only refines node representations but also amplifies reconstruction errors for anomalies, thereby boosting the model’s detection performance across both injected and ground-truth datasets.

\begin{table*}[t!]
 \centering
 \scalebox{1.0}{
 \begin{tabular}{c|ccccccc}
\toprule
Method & Citeseer   & ACM        & Blogcatalog & Weibo      & Reddit     & YelpRes     & Amazon     \\ \midrule
DOMINANT \cite{ding2019deep}  & 71.70 ± 0.27 & 79.38 ± 1.77 & \underline{78.36 ± 0.15}  & 90.51 ± 0.06 & 50.82 ± 0.14 & \underline{77.03 ± 0.12}  & \underline{88.71 ± 2.29} \\
ANEMONE \cite{jin2021anemone}  & 89.75 ± 2.47 & 82.63 ± 1.83 & 56.23 ± 0.78  & 28.86 ± 1.26 & \underline{57.99 ± 0.99} & 48.73 ± 7.99  & 78.72 ±1.42 \\
CoLA \cite{liu2021anomaly}     & 88.47 ± 0.80 & 78.90 ± 1.47 & 60.75 ± 0.56  & 28.12 ± 1.06 & 53.59 ± 7.81 & 63.63 ± 11.51 & 76.85 ± 1.14 \\
SL-GAD \cite{zheng2021generative}   & \underline{93.24 ± 0.92} & 81.50 ± 0.65 & 70.16 ± 0.88  & 74.93 ± 0.21 & 54.73 ± 0.61 & 73.19 ± 0.49  & 87.73 ± 0.96 \\
GRADATE \cite{duan2023graph}   & 83.83 ± 0.47 & \underline{85.26 ± 1.09} & 58.30 ± 0.51  & 31.87 ± 2.24 & 48.84 ± 1.51 & 47.01 ± 10.41 & 75.10 ± 1.20 \\
GADAM  \cite{zhou2024graph}   & 62.53 ± 0.43 & 70.13 ± 0.42 & 67.99 ± 0.45  & 27.63 ± 0.60 & 57.81 ± 0.53 & 17.02 ± 1.62  & 58.14 ± 1.09 \\
ADA-GAD  \cite{he2024ada}   & 87.02 ± 0.01 & 73.33 ± 0.01 & 69.46 ± 0.02  & \underline{92.50 ± 0.48} & 56.89 ± 0.01 & 60.15 ± 0.32  & 84.23 ± 0.53 \\
DiffGAD  \cite{li2024diffgad}   & 90.25 ± 1.34 & 76.42 ± 0.08 & 67.57 ± 0.04  & \textbf{93.40 ± 0.30} & 56.30 ± 0.10 & 75.14 ± 0.45  & 87.88 ± 1.21 \\
GAD-NR \cite{roy2024gad}   & 72.00 ± 0.21 & 72.75 ± 0.12 & 57.05 ± 0.29  & 76.68 ± 0.65 & 57.99 ± 1.67 & 56.35 ± 0.26  & 81.42 ± 0.11 \\ \midrule
GTHNA    & \textbf{93.62 ± 0.12} & \textbf{89.81 ± 0.19} & \textbf{80.23 ± 0.12}  & 92.38 ± 0.15 & \textbf{59.21 ± 1.63} & \textbf{86.88 ± 0.46}  & \textbf{90.41 ± 0.53} \\ \bottomrule
\end{tabular}
 }
\caption{AUC values (\%) on seven datasets.}
\label{tab:2}
\end{table*}

\subsection{Ablation Studies (RQ2)}
To gain deeper insights into the impact of the structure extractor and the memory module on model performance, we conduct ablation experiment targeting these two components. Specifically, we define the following variants:
\begin{itemize} 
\item [$\bullet$] \textbf{GTHNA-S}: This variant excludes the structure extractor and relies solely on the Laplacian matrix to generate positional encodings. 
\item [$\bullet$] \textbf{GTHNA-M}: This variant removes the memory module that records normal node patterns, and directly uses node embeddings generated by the encoder to perform multi-scale node reconstruction. 
\end{itemize}

\begin{figure}[h!]
    \centering
    \includegraphics[width=1.0\linewidth]{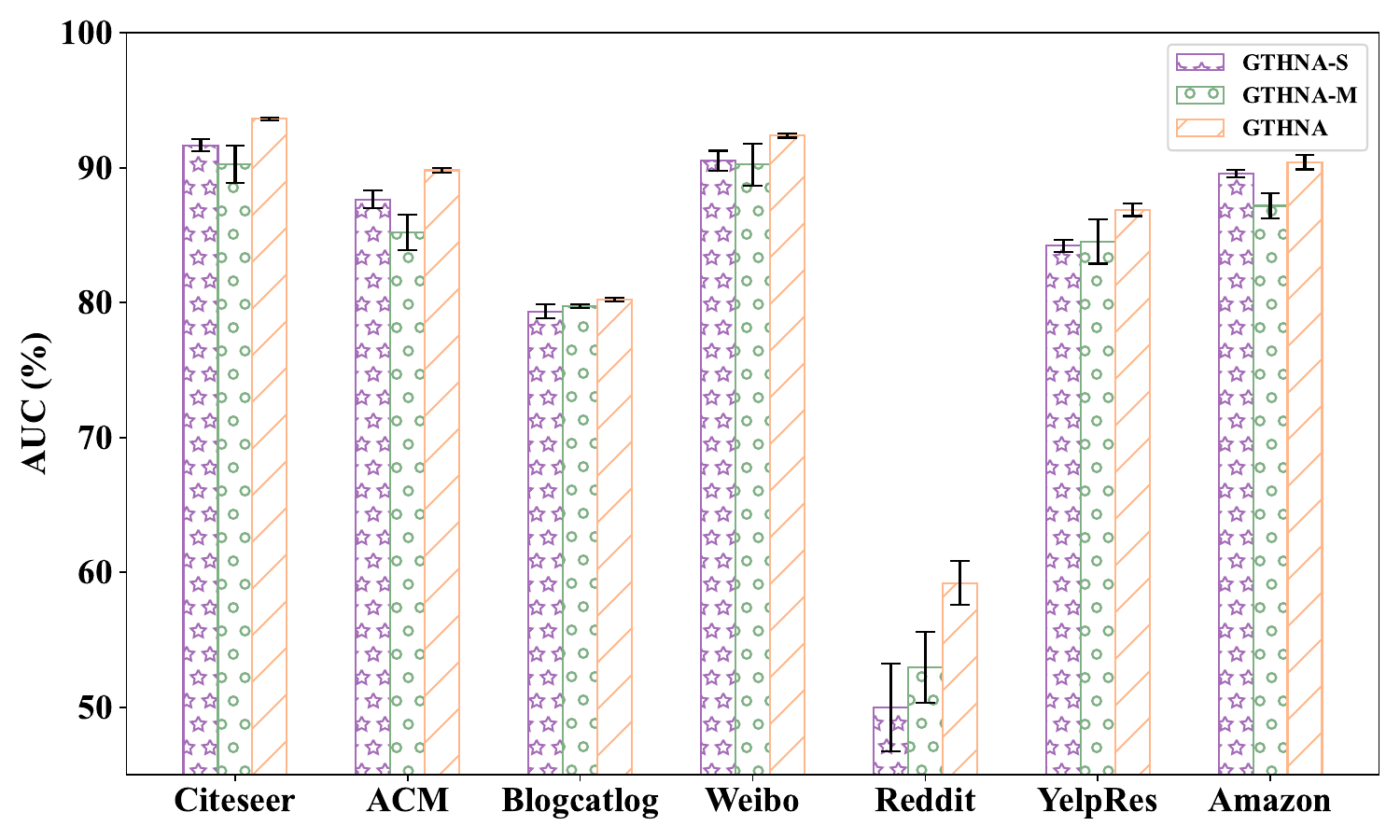}
    \caption{Comparison of ablation experiment results.}
    \label{fig:4}
\end{figure}

As shown in \figurename~\ref{fig:4}, the full GTHNA model achieves the best performance, demonstrating the effectiveness of both the structure extractor and the memory module. Furthermore, it can be observed that the memory module plays a more crucial role in model performance across most datasets, particularly those with injected anomalies. This is because, in these injected datasets, the anomaly noise is more pronounced, significantly affecting the graph reconstruction process.

\subsection{Analysis of the Memory Module (RQ3)}
\noindent To investigate the ability of the memory module to effectively capture and store the patterns of normal nodes, we perform experiment on both the injected dataset Citeseer and the ground-truth dataset Weibo. 

\begin{figure}[h!]
    \centering
    \includegraphics[width=1.0\linewidth]{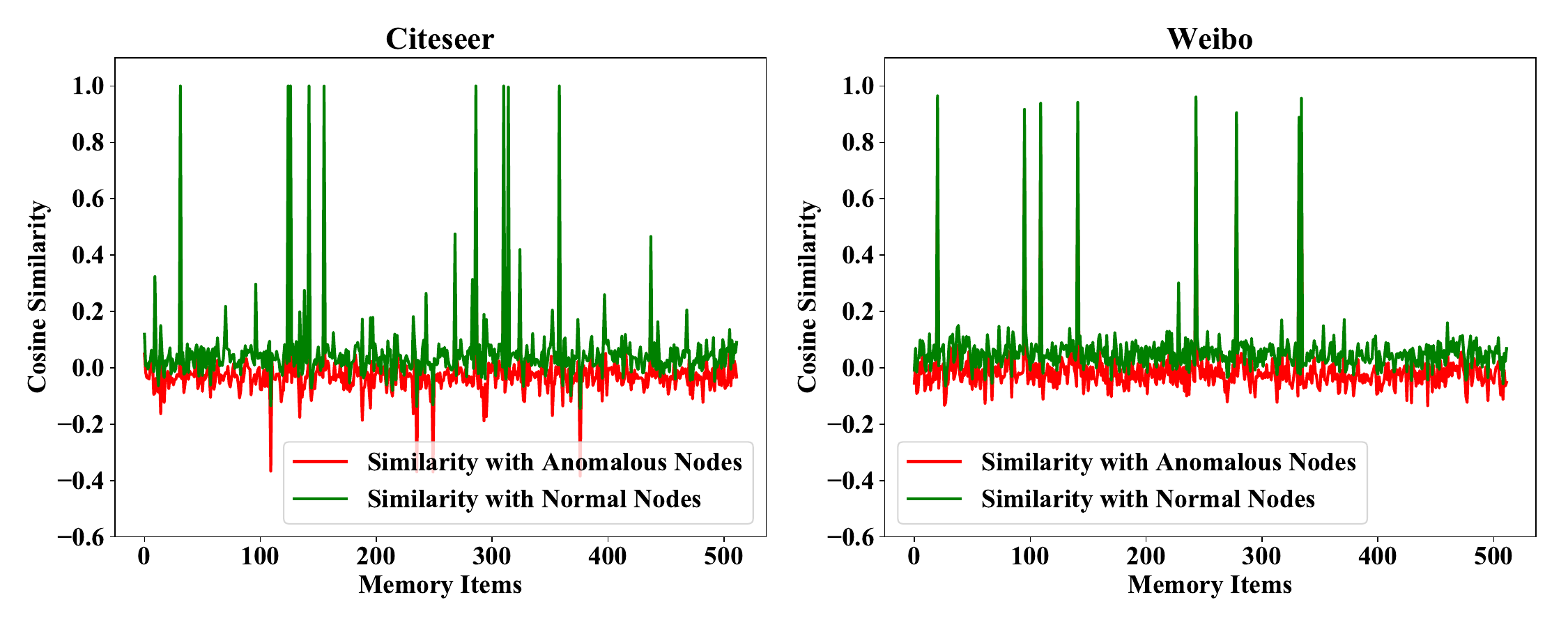}
    \caption{Cosine similarity comparison between memory items and normal/anomalous nodes on Citeseer and Weibo.}
    \label{fig:55}
\end{figure}
As shown in \figurename~\ref{fig:55}, the similarity distributions reveal that the memory items in GTHNA consistently exhibit higher alignment with normal nodes and lower similarity with anomalous ones. This clear separation highlights the memory module’s strong discriminative capability. Specifically, memory items are updated in an adaptive manner by aggregating information from nodes with low reconstruction error, which are more likely to be normal. Through this selective update mechanism, the memory effectively evolves into a stable and compact repository of normal behavior patterns. By integrating this memory-guided process into the anomaly evaluation pipeline, GTHNA enhances its robustness against anomaly contamination during learning.

\subsection{Anomaly Score Analysis (RQ4)}
To assess the distinguishability of node anomaly scores computed by GTHNA, we visualize the anomaly score distributions on 3 datasets. Notably, the number of normal nodes significantly surpasses that of anomalous nodes in most datasets. To avoid disproportionate representation, we visualize the anomaly scores of all anomalous nodes and randomly select a subset of normal nodes, totaling 5 times the number of anomalous nodes for comparison.

\begin{figure}[h!]
    \centering
    \includegraphics[width=1.0\linewidth]{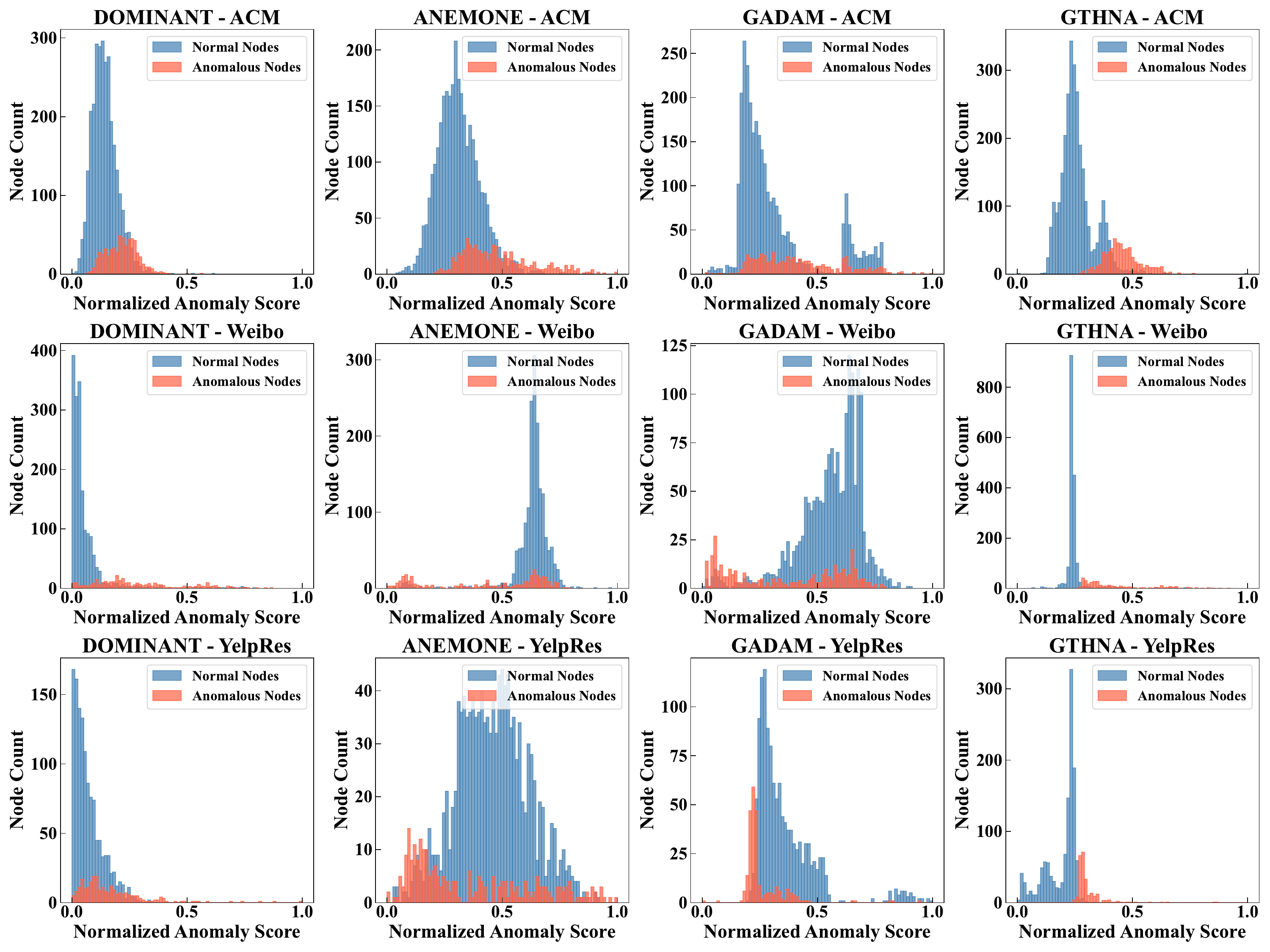}
    \caption{Visualization of anomaly scores.}
    \label{fig:66}
\end{figure}

Compared with other methods in \figurename~\ref{fig:66}, we can observe that GTHNA can effectively increase the gap between the anomaly scores of normal and anomalous nodes. GTHNA demonstrates a significantly clearer separation between normal and anomalous nodes compared to existing methods, with anomalous nodes consistently assigned higher scores. This enhanced discriminative capability stems from the holistic design of GTHNA, which integrates a structure-aware local-global Transformer encoder, a memory-guided reconstruction mechanism, and a multi-scale anomaly evaluation strategy. Together, these components enable the model to generate more expressive and robust node representations, suppress the influence of outliers during reconstruction, and capture anomalies at various structural levels. The resulting anomaly scores are more distinguishable, confirming the effectiveness of our framework in addressing key challenges such as over-smoothing and anomalous interference in graph anomaly detection.

\subsection{Hyperparameter Analysis (RQ5)}
\begin{figure}[htbp]
    \centering
    \includegraphics[width=1.0\linewidth]{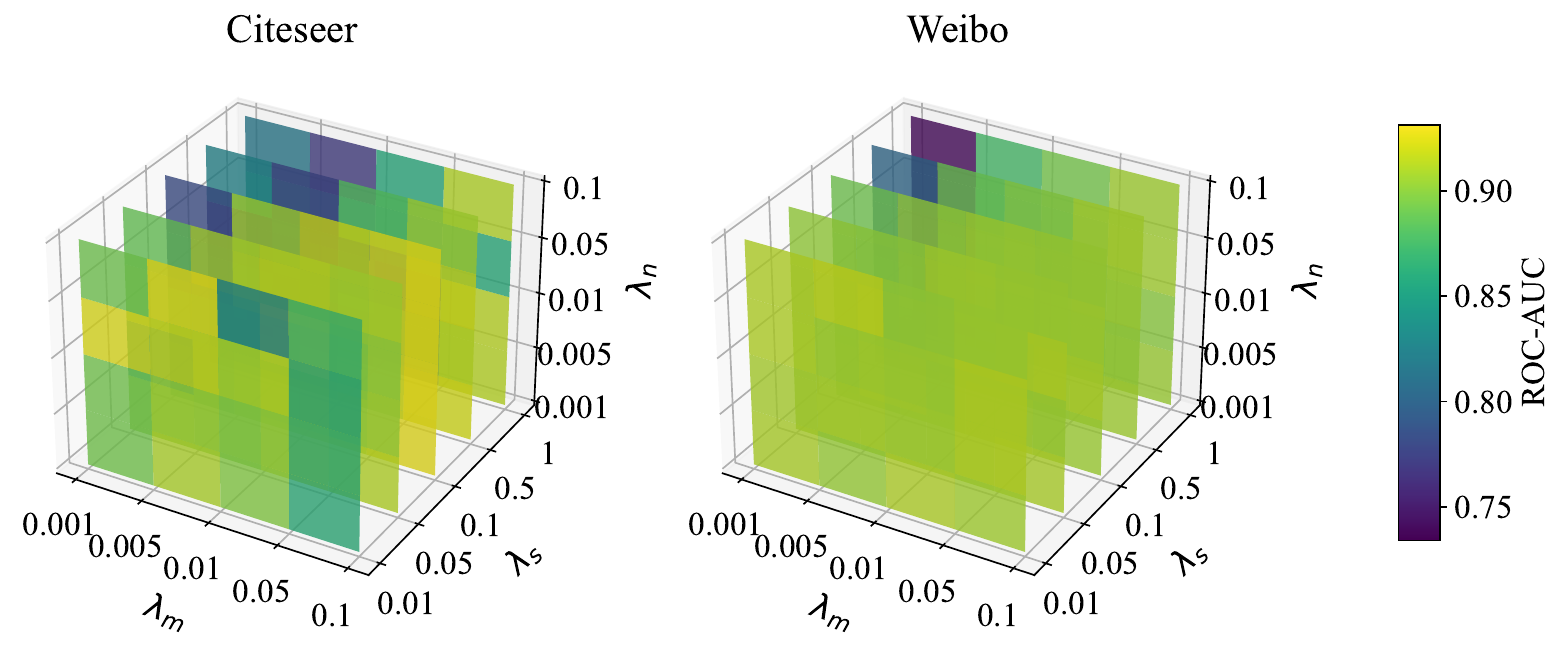}
    \caption{Impacts of loss weights $\lambda_s$, $\lambda_n$ and $\lambda_m$ on model performance.}
    \label{fig:hyper}
\end{figure}

We further analyze the trend of GTHNA's performance under different loss weight settings ($\lambda_s$, $\lambda_n$, and $\lambda_m$) on Citeseer and Weibo. The random seed is set to 0 in this experiment. As shown in \figurename~\ref{fig:hyper}, the performance on Weibo is relatively insensitive to changes in hyperparameters. In most configurations, our model demonstrates consistent anomaly detection performance. Meanwhile, on the injected dataset, Citeseer, the performance increases with higher values of $\lambda_m$, suggesting that the injected datasets rely more heavily on the memory module to record normal node patterns, thereby improving the model's ability to distinguish anomalies.

\section{Conclusion and Outlook}

To address the over-smoothing issue in node representation learning with GCNs and the interference of anomalous nodes in graph reconstruction, we propose a holistic memory-Transformer framework for anomaly detection (GTHNA). By enhancing the node positional encoding, our method improves node representation distinctiveness, allowing for more accurate differentiation between normal and anomalous nodes. Additionally, we introduce the memory module to record normal node patterns and guide the graph reconstruction process, effectively mitigating the impact of anomalous nodes during training. Experimental results show that our method outperforms existing approaches on seven benchmark datasets, demonstrating its robustness and effectiveness in various settings. Future work could focus on optimizing the computation of positional encodings for large-scale graphs to reduce computational cost while maintaining performance. Additionally, exploring adaptive memory mechanisms could improve the handling of noisy data and enhance the model's generalizability to unseen anomaly patterns.

\begin{thebibliography}{42}


\ifx \showCODEN    \undefined \def \showCODEN     #1{\unskip}     \fi
\ifx \showISBNx    \undefined \def \showISBNx     #1{\unskip}     \fi
\ifx \showISBNxiii \undefined \def \showISBNxiii  #1{\unskip}     \fi
\ifx \showISSN     \undefined \def \showISSN      #1{\unskip}     \fi
\ifx \showLCCN     \undefined \def \showLCCN      #1{\unskip}     \fi
\ifx \shownote     \undefined \def \shownote      #1{#1}          \fi
\ifx \showarticletitle \undefined \def \showarticletitle #1{#1}   \fi
\ifx \showURL      \undefined \def \showURL       {\relax}        \fi
\providecommand\bibfield[2]{#2}
\providecommand\bibinfo[2]{#2}
\providecommand\natexlab[1]{#1}
\providecommand\showeprint[2][]{arXiv:#2}

\bibitem[Chen et~al\mbox{.}(2020)]%
        {chen2020measuring}
\bibfield{author}{\bibinfo{person}{Deli Chen}, \bibinfo{person}{Yankai Lin}, \bibinfo{person}{Wei Li}, \bibinfo{person}{Peng Li}, \bibinfo{person}{Jie Zhou}, {and} \bibinfo{person}{Xu Sun}.} \bibinfo{year}{2020}\natexlab{}.
\newblock \showarticletitle{Measuring and relieving the over-smoothing problem for graph neural networks from the topological view}. In \bibinfo{booktitle}{\emph{AAAI}}, Vol.~\bibinfo{volume}{34}. \bibinfo{pages}{3438--3445}.
\newblock


\bibitem[Chen et~al\mbox{.}(2022)]%
        {chen2022structure}
\bibfield{author}{\bibinfo{person}{Dexiong Chen}, \bibinfo{person}{Leslie O’Bray}, {and} \bibinfo{person}{Karsten Borgwardt}.} \bibinfo{year}{2022}\natexlab{}.
\newblock \showarticletitle{Structure-aware transformer for graph representation learning}. In \bibinfo{booktitle}{\emph{ICML}}. PMLR, \bibinfo{pages}{3469--3489}.
\newblock


\bibitem[Ding et~al\mbox{.}(2019)]%
        {ding2019deep}
\bibfield{author}{\bibinfo{person}{Kaize Ding}, \bibinfo{person}{Jundong Li}, \bibinfo{person}{Rohit Bhanushali}, {and} \bibinfo{person}{Huan Liu}.} \bibinfo{year}{2019}\natexlab{}.
\newblock \showarticletitle{Deep anomaly detection on attributed networks}. In \bibinfo{booktitle}{\emph{SDM}}. SIAM, \bibinfo{pages}{594--602}.
\newblock


\bibitem[Du et~al\mbox{.}(2022)]%
        {du2022graph}
\bibfield{author}{\bibinfo{person}{Xusheng Du}, \bibinfo{person}{Jiong Yu}, \bibinfo{person}{Zheng Chu}, \bibinfo{person}{Lina Jin}, {and} \bibinfo{person}{Jiaying Chen}.} \bibinfo{year}{2022}\natexlab{}.
\newblock \showarticletitle{Graph autoencoder-based unsupervised outlier detection}.
\newblock \bibinfo{journal}{\emph{Information Sciences}}  \bibinfo{volume}{608} (\bibinfo{year}{2022}), \bibinfo{pages}{532--550}.
\newblock


\bibitem[Duan et~al\mbox{.}(2023)]%
        {duan2023graph}
\bibfield{author}{\bibinfo{person}{Jingcan Duan}, \bibinfo{person}{Siwei Wang}, \bibinfo{person}{Pei Zhang}, \bibinfo{person}{En Zhu}, \bibinfo{person}{Jingtao Hu}, \bibinfo{person}{Hu Jin}, \bibinfo{person}{Yue Liu}, {and} \bibinfo{person}{Zhibin Dong}.} \bibinfo{year}{2023}\natexlab{}.
\newblock \showarticletitle{Graph anomaly detection via multi-scale contrastive learning networks with augmented view}. In \bibinfo{booktitle}{\emph{AAAI}}, Vol.~\bibinfo{volume}{37}. \bibinfo{pages}{7459--7467}.
\newblock


\bibitem[Fan et~al\mbox{.}(2021)]%
        {fan2021hyperspectral}
\bibfield{author}{\bibinfo{person}{Ganghui Fan}, \bibinfo{person}{Yong Ma}, \bibinfo{person}{Xiaoguang Mei}, \bibinfo{person}{Fan Fan}, \bibinfo{person}{Jun Huang}, {and} \bibinfo{person}{Jiayi Ma}.} \bibinfo{year}{2021}\natexlab{}.
\newblock \showarticletitle{Hyperspectral anomaly detection with robust graph autoencoders}.
\newblock \bibinfo{journal}{\emph{IEEE Transactions on Geoscience and Remote Sensing}}  \bibinfo{volume}{60} (\bibinfo{year}{2021}), \bibinfo{pages}{1--14}.
\newblock


\bibitem[Fan et~al\mbox{.}(2020)]%
        {fan2020anomalydae}
\bibfield{author}{\bibinfo{person}{Haoyi Fan}, \bibinfo{person}{Fengbin Zhang}, {and} \bibinfo{person}{Zuoyong Li}.} \bibinfo{year}{2020}\natexlab{}.
\newblock \showarticletitle{Anomalydae: Dual autoencoder for anomaly detection on attributed networks}. In \bibinfo{booktitle}{\emph{ICASSP}}. IEEE, \bibinfo{pages}{5685--5689}.
\newblock


\bibitem[Guo et~al\mbox{.}(2024)]%
        {guo2024transgad}
\bibfield{author}{\bibinfo{person}{Zehao Guo}, \bibinfo{person}{Nannan Wu}, \bibinfo{person}{Yiming Zhao}, {and} \bibinfo{person}{Wenjun Wang}.} \bibinfo{year}{2024}\natexlab{}.
\newblock \showarticletitle{TransGAD: A Transformer-Based Autoencoder for Graph Anomaly Detection}. In \bibinfo{booktitle}{\emph{DASFAA}}. Springer, \bibinfo{pages}{269--284}.
\newblock


\bibitem[He et~al\mbox{.}(2024)]%
        {he2024ada}
\bibfield{author}{\bibinfo{person}{Junwei He}, \bibinfo{person}{Qianqian Xu}, \bibinfo{person}{Yangbangyan Jiang}, \bibinfo{person}{Zitai Wang}, {and} \bibinfo{person}{Qingming Huang}.} \bibinfo{year}{2024}\natexlab{}.
\newblock \showarticletitle{Ada-gad: Anomaly-denoised autoencoders for graph anomaly detection}. In \bibinfo{booktitle}{\emph{AAAI}}, Vol.~\bibinfo{volume}{38}. \bibinfo{pages}{8481--8489}.
\newblock


\bibitem[Huang et~al\mbox{.}(2023)]%
        {huang2023unsupervised}
\bibfield{author}{\bibinfo{person}{Yihong Huang}, \bibinfo{person}{Liping Wang}, \bibinfo{person}{Fan Zhang}, {and} \bibinfo{person}{Xuemin Lin}.} \bibinfo{year}{2023}\natexlab{}.
\newblock \showarticletitle{Unsupervised graph outlier detection: Problem revisit, new insight, and superior method}. In \bibinfo{booktitle}{\emph{ICDE}}. IEEE, \bibinfo{pages}{2565--2578}.
\newblock


\bibitem[Hussain et~al\mbox{.}(2021)]%
        {hussain2021edge}
\bibfield{author}{\bibinfo{person}{Md~Shamim Hussain}, \bibinfo{person}{Mohammed~J Zaki}, {and} \bibinfo{person}{Dharmashankar Subramanian}.} \bibinfo{year}{2021}\natexlab{}.
\newblock \showarticletitle{Edge-augmented graph transformers: Global self-attention is enough for graphs}.
\newblock \bibinfo{journal}{\emph{arXiv preprint arXiv:2108.03348}}  \bibinfo{volume}{3} (\bibinfo{year}{2021}).
\newblock


\bibitem[Jin et~al\mbox{.}(2021)]%
        {jin2021anemone}
\bibfield{author}{\bibinfo{person}{Ming Jin}, \bibinfo{person}{Yixin Liu}, \bibinfo{person}{Yu Zheng}, \bibinfo{person}{Lianhua Chi}, \bibinfo{person}{Yuan-Fang Li}, {and} \bibinfo{person}{Shirui Pan}.} \bibinfo{year}{2021}\natexlab{}.
\newblock \showarticletitle{Anemone: Graph anomaly detection with multi-scale contrastive learning}. In \bibinfo{booktitle}{\emph{CIKM}}. \bibinfo{pages}{3122--3126}.
\newblock


\bibitem[Kaghazgaran et~al\mbox{.}(2018)]%
        {kaghazgaran2018combating}
\bibfield{author}{\bibinfo{person}{Parisa Kaghazgaran}, \bibinfo{person}{James Caverlee}, {and} \bibinfo{person}{Anna Squicciarini}.} \bibinfo{year}{2018}\natexlab{}.
\newblock \showarticletitle{Combating crowdsourced review manipulators: A neighborhood-based approach}. In \bibinfo{booktitle}{\emph{WSDM}}. \bibinfo{pages}{306--314}.
\newblock


\bibitem[Kipf and Welling(2016)]%
        {kipf2016semi}
\bibfield{author}{\bibinfo{person}{Thomas~N Kipf} {and} \bibinfo{person}{Max Welling}.} \bibinfo{year}{2016}\natexlab{}.
\newblock \showarticletitle{Semi-supervised classification with graph convolutional networks}.
\newblock \bibinfo{journal}{\emph{arXiv preprint arXiv:1609.02907}} (\bibinfo{year}{2016}).
\newblock


\bibitem[Kumar et~al\mbox{.}(2019)]%
        {kumar2019predicting}
\bibfield{author}{\bibinfo{person}{Srijan Kumar}, \bibinfo{person}{Xikun Zhang}, {and} \bibinfo{person}{Jure Leskovec}.} \bibinfo{year}{2019}\natexlab{}.
\newblock \showarticletitle{Predicting dynamic embedding trajectory in temporal interaction networks}. In \bibinfo{booktitle}{\emph{SIGKDD}}. \bibinfo{pages}{1269--1278}.
\newblock


\bibitem[Kurshan et~al\mbox{.}(2020)]%
        {kurshan2020financial}
\bibfield{author}{\bibinfo{person}{Eren Kurshan}, \bibinfo{person}{Hongda Shen}, {and} \bibinfo{person}{Haojie Yu}.} \bibinfo{year}{2020}\natexlab{}.
\newblock \showarticletitle{Financial crime \& fraud detection using graph computing: Application considerations \& outlook}. In \bibinfo{booktitle}{\emph{TransAI}}. IEEE, \bibinfo{pages}{125--130}.
\newblock


\bibitem[Li et~al\mbox{.}(2024)]%
        {li2024diffgad}
\bibfield{author}{\bibinfo{person}{Jinghan Li}, \bibinfo{person}{Yuan Gao}, \bibinfo{person}{Jinda Lu}, \bibinfo{person}{Junfeng Fang}, \bibinfo{person}{Congcong Wen}, \bibinfo{person}{Hui Lin}, {and} \bibinfo{person}{Xiang Wang}.} \bibinfo{year}{2024}\natexlab{}.
\newblock \showarticletitle{Diffgad: A diffusion-based unsupervised graph anomaly detector}.
\newblock \bibinfo{journal}{\emph{arXiv preprint arXiv:2410.06549}} (\bibinfo{year}{2024}).
\newblock



\bibitem[Liu et~al\mbox{.}(2021)]%
        {liu2021anomaly}
\bibfield{author}{\bibinfo{person}{Yixin Liu}, \bibinfo{person}{Zhao Li}, \bibinfo{person}{Shirui Pan}, \bibinfo{person}{Chen Gong}, \bibinfo{person}{Chuan Zhou}, {and} \bibinfo{person}{George Karypis}.} \bibinfo{year}{2021}\natexlab{}.
\newblock \showarticletitle{Anomaly detection on attributed networks via contrastive self-supervised learning}.
\newblock \bibinfo{journal}{\emph{IEEE Transactions on Neural Networks and Learning Systems}} \bibinfo{volume}{33}, \bibinfo{number}{6} (\bibinfo{year}{2021}), \bibinfo{pages}{2378--2392}.
\newblock


\bibitem[Lu et~al\mbox{.}(2024)]%
        {lu2024enhancing}
\bibfield{author}{\bibinfo{person}{Qingcheng Lu}, \bibinfo{person}{Nannan Wu}, \bibinfo{person}{Yiming Zhao}, \bibinfo{person}{Wenjun Wang}, {and} \bibinfo{person}{Quannan Zu}.} \bibinfo{year}{2024}\natexlab{}.
\newblock \showarticletitle{Enhancing Multi-view Contrastive Learning for Graph Anomaly Detection}. In \bibinfo{booktitle}{\emph{DASFAA}}. Springer, \bibinfo{pages}{236--251}.
\newblock


\bibitem[Luo et~al\mbox{.}(2022)]%
        {luo2022comga}
\bibfield{author}{\bibinfo{person}{Xuexiong Luo}, \bibinfo{person}{Jia Wu}, \bibinfo{person}{Amin Beheshti}, \bibinfo{person}{Jian Yang}, \bibinfo{person}{Xiankun Zhang}, \bibinfo{person}{Yuan Wang}, {and} \bibinfo{person}{Shan Xue}.} \bibinfo{year}{2022}\natexlab{}.
\newblock \showarticletitle{Comga: Community-aware attributed graph anomaly detection}. In \bibinfo{booktitle}{\emph{WSDM}}. \bibinfo{pages}{657--665}.
\newblock


\bibitem[Ma et~al\mbox{.}(2021)]%
        {ma2021comprehensive}
\bibfield{author}{\bibinfo{person}{Xiaoxiao Ma}, \bibinfo{person}{Jia Wu}, \bibinfo{person}{Shan Xue}, \bibinfo{person}{Jian Yang}, \bibinfo{person}{Chuan Zhou}, \bibinfo{person}{Quan~Z Sheng}, \bibinfo{person}{Hui Xiong}, {and} \bibinfo{person}{Leman Akoglu}.} \bibinfo{year}{2021}\natexlab{}.
\newblock \showarticletitle{A comprehensive survey on graph anomaly detection with deep learning}.
\newblock \bibinfo{journal}{\emph{IEEE Transactions on Knowledge and Data Engineering}} \bibinfo{volume}{35}, \bibinfo{number}{12} (\bibinfo{year}{2021}), \bibinfo{pages}{12012--12038}.
\newblock


\bibitem[Park et~al\mbox{.}(2020)]%
        {park2020learning}
\bibfield{author}{\bibinfo{person}{Hyunjong Park}, \bibinfo{person}{Jongyoun Noh}, {and} \bibinfo{person}{Bumsub Ham}.} \bibinfo{year}{2020}\natexlab{}.
\newblock \showarticletitle{Learning memory-guided normality for anomaly detection}. In \bibinfo{booktitle}{\emph{CVPR}}. \bibinfo{pages}{14372--14381}.
\newblock


\bibitem[Rao et~al\mbox{.}(2021)]%
        {rao2021review}
\bibfield{author}{\bibinfo{person}{Sanjeev Rao}, \bibinfo{person}{Anil~Kumar Verma}, {and} \bibinfo{person}{Tarunpreet Bhatia}.} \bibinfo{year}{2021}\natexlab{}.
\newblock \showarticletitle{A review on social spam detection: Challenges, open issues, and future directions}.
\newblock \bibinfo{journal}{\emph{Expert Systems with Applications}}  \bibinfo{volume}{186} (\bibinfo{year}{2021}), \bibinfo{pages}{115742}.
\newblock


\bibitem[Rayana and Akoglu(2015)]%
        {rayana2015collective}
\bibfield{author}{\bibinfo{person}{Shebuti Rayana} {and} \bibinfo{person}{Leman Akoglu}.} \bibinfo{year}{2015}\natexlab{}.
\newblock \showarticletitle{Collective opinion spam detection: Bridging review networks and metadata}. In \bibinfo{booktitle}{\emph{SIGKDD}}. \bibinfo{pages}{985--994}.
\newblock


\bibitem[Roy et~al\mbox{.}(2024)]%
        {roy2024gad}
\bibfield{author}{\bibinfo{person}{Amit Roy}, \bibinfo{person}{Juan Shu}, \bibinfo{person}{Jia Li}, \bibinfo{person}{Carl Yang}, \bibinfo{person}{Olivier Elshocht}, \bibinfo{person}{Jeroen Smeets}, {and} \bibinfo{person}{Pan Li}.} \bibinfo{year}{2024}\natexlab{}.
\newblock \showarticletitle{Gad-nr: Graph anomaly detection via neighborhood reconstruction}. In \bibinfo{booktitle}{\emph{WSDM}}. \bibinfo{pages}{576--585}.
\newblock


\bibitem[Sen et~al\mbox{.}(2008)]%
        {sen2008collective}
\bibfield{author}{\bibinfo{person}{Prithviraj Sen}, \bibinfo{person}{Galileo Namata}, \bibinfo{person}{Mustafa Bilgic}, \bibinfo{person}{Lise Getoor}, \bibinfo{person}{Brian Galligher}, {and} \bibinfo{person}{Tina Eliassi-Rad}.} \bibinfo{year}{2008}\natexlab{}.
\newblock \showarticletitle{Collective classification in network data}.
\newblock \bibinfo{journal}{\emph{AI Magazine}} \bibinfo{volume}{29}, \bibinfo{number}{3} (\bibinfo{year}{2008}), \bibinfo{pages}{93--93}.
\newblock



\bibitem[Shaw et~al\mbox{.}(2018)]%
        {shaw2018self}
\bibfield{author}{\bibinfo{person}{Peter Shaw}, \bibinfo{person}{Jakob Uszkoreit}, {and} \bibinfo{person}{Ashish Vaswani}.} \bibinfo{year}{2018}\natexlab{}.
\newblock \showarticletitle{Self-attention with relative position representations}.
\newblock \bibinfo{journal}{\emph{arXiv preprint arXiv:1803.02155}} (\bibinfo{year}{2018}).
\newblock


\bibitem[Tang et~al\mbox{.}(2022)]%
        {tang2022rethinking}
\bibfield{author}{\bibinfo{person}{Jianheng Tang}, \bibinfo{person}{Jiajin Li}, \bibinfo{person}{Ziqi Gao}, {and} \bibinfo{person}{Jia Li}.} \bibinfo{year}{2022}\natexlab{}.
\newblock \showarticletitle{Rethinking graph neural networks for anomaly detection}. In \bibinfo{booktitle}{\emph{ICML}}. PMLR, \bibinfo{pages}{21076--21089}.
\newblock


\bibitem[Tang et~al\mbox{.}(2008)]%
        {tang2008arnetminer}
\bibfield{author}{\bibinfo{person}{Jie Tang}, \bibinfo{person}{Jing Zhang}, \bibinfo{person}{Limin Yao}, \bibinfo{person}{Juanzi Li}, \bibinfo{person}{Li Zhang}, {and} \bibinfo{person}{Zhong Su}.} \bibinfo{year}{2008}\natexlab{}.
\newblock \showarticletitle{Arnetminer: extraction and mining of academic social networks}. In \bibinfo{booktitle}{\emph{SIGKDD}}. \bibinfo{pages}{990--998}.
\newblock


\bibitem[Tang and Liu(2009)]%
        {tang2009relational}
\bibfield{author}{\bibinfo{person}{Lei Tang} {and} \bibinfo{person}{Huan Liu}.} \bibinfo{year}{2009}\natexlab{}.
\newblock \showarticletitle{Relational learning via latent social dimensions}. In \bibinfo{booktitle}{\emph{SIGKDD}}. \bibinfo{pages}{817--826}.
\newblock


\bibitem[Veli{\v{c}}kovi{\'c}(2023)]%
        {velivckovic2023everything}
\bibfield{author}{\bibinfo{person}{Petar Veli{\v{c}}kovi{\'c}}.} \bibinfo{year}{2023}\natexlab{}.
\newblock \showarticletitle{Everything is connected: Graph neural networks}.
\newblock \bibinfo{journal}{\emph{Current Opinion in Structural Biology}}  \bibinfo{volume}{79} (\bibinfo{year}{2023}), \bibinfo{pages}{102538}.
\newblock


\bibitem[Wu et~al\mbox{.}(2024)]%
        {wu2024demystifying}
\bibfield{author}{\bibinfo{person}{Xinyi Wu}, \bibinfo{person}{Amir Ajorlou}, \bibinfo{person}{Zihui Wu}, {and} \bibinfo{person}{Ali Jadbabaie}.} \bibinfo{year}{2024}\natexlab{}.
\newblock \showarticletitle{Demystifying oversmoothing in attention-based graph neural networks}.
\newblock \bibinfo{journal}{\emph{Advances in Neural Information Processing Systems}}  \bibinfo{volume}{36} (\bibinfo{year}{2024}).
\newblock


\bibitem[Wu et~al\mbox{.}(2020)]%
        {wu2020network}
\bibfield{author}{\bibinfo{person}{Yirui Wu}, \bibinfo{person}{Dabao Wei}, {and} \bibinfo{person}{Jun Feng}.} \bibinfo{year}{2020}\natexlab{}.
\newblock \showarticletitle{Network attacks detection methods based on deep learning techniques: a survey}.
\newblock \bibinfo{journal}{\emph{Security and Communication Networks}} \bibinfo{volume}{2020}, \bibinfo{number}{1} (\bibinfo{year}{2020}), \bibinfo{pages}{8872923}.
\newblock

\bibitem[Zhang et~al\mbox{.}(2020)]%
        {zhang2020graph}
\bibfield{author}{\bibinfo{person}{Jiawei Zhang}, \bibinfo{person}{Haopeng Zhang}, \bibinfo{person}{Congying Xia}, {and} \bibinfo{person}{Li Sun}.} \bibinfo{year}{2020}\natexlab{}.
\newblock \showarticletitle{Graph-bert: Only attention is needed for learning graph representations}.
\newblock \bibinfo{journal}{\emph{arXiv preprint arXiv:2001.05140}} (\bibinfo{year}{2020}).
\newblock


\bibitem[Zhao et~al\mbox{.}(2020)]%
        {zhao2020error}
\bibfield{author}{\bibinfo{person}{Tong Zhao}, \bibinfo{person}{Chuchen Deng}, \bibinfo{person}{Kaifeng Yu}, \bibinfo{person}{Tianwen Jiang}, \bibinfo{person}{Daheng Wang}, {and} \bibinfo{person}{Meng Jiang}.} \bibinfo{year}{2020}\natexlab{}.
\newblock \showarticletitle{Error-bounded graph anomaly loss for gnns}. In \bibinfo{booktitle}{\emph{CIKM}}. \bibinfo{pages}{1873--1882}.
\newblock


\bibitem[Zheng et~al\mbox{.}(2021)]%
        {zheng2021generative}
\bibfield{author}{\bibinfo{person}{Yu Zheng}, \bibinfo{person}{Ming Jin}, \bibinfo{person}{Yixin Liu}, \bibinfo{person}{Lianhua Chi}, \bibinfo{person}{Khoa~T Phan}, {and} \bibinfo{person}{Yi-Ping~Phoebe Chen}.} \bibinfo{year}{2021}\natexlab{}.
\newblock \showarticletitle{Generative and contrastive self-supervised learning for graph anomaly detection}.
\newblock \bibinfo{journal}{\emph{IEEE Transactions on Knowledge and Data Engineering}} \bibinfo{volume}{35}, \bibinfo{number}{12} (\bibinfo{year}{2021}), \bibinfo{pages}{12220--12233}.
\newblock


\bibitem[Zhou et~al\mbox{.}(2024)]%
        {zhou2024graph}
\bibfield{author}{\bibinfo{person}{Qinghai Zhou}, \bibinfo{person}{Yuzhong Chen}, \bibinfo{person}{Zhe Xu}, \bibinfo{person}{Yuhang Wu}, \bibinfo{person}{Menghai Pan}, \bibinfo{person}{Mahashweta Das}, \bibinfo{person}{Hao Yang}, {and} \bibinfo{person}{Hanghang Tong}.} \bibinfo{year}{2024}\natexlab{}.
\newblock \showarticletitle{Graph Anomaly Detection with Adaptive Node Mixup}. In \bibinfo{booktitle}{\emph{CIKM}}. \bibinfo{pages}{3494--3504}.
\newblock


\end{thebibliography}

\bibliographystyle{ACM-Reference-Format}
\balance


\end{document}